# A Survey on Deep Neural Networks in Collaborative Filtering Recommendation Systems

Pang Li [1], Shahrul Azman Mohd Noah [2], Hafiz Mohd Sarim [3]

Center for Artificial Intelligence and Technology

Faculty of Information Science and Technology

The National University of Malaysia

a184459@siswa.ukm.edu.my[1], shahrul@ukm.edu.my[2], hms@ukm.edu.my[3]

**ABSTRACT**

This survey provides a examination of the use of Deep Neural Networks (DNN) in Collaborative Filtering (CF) recommendation systems. As the digital world increasingly relies on data-driven approaches, traditional CF techniques face limitations in scalability and flexibility. DNNs can address these challenges by effectively modeling complex, non-linear relationships within the data. We begin by exploring the fundamental principles of both collaborative filtering and deep neural networks, laying the groundwork for understanding their integration. Subsequently, we review key advancements in the field, categorizing various deep learning models that enhance CF systems, including Multilayer Perceptrons (MLP), Convolutional Neural Networks (CNN), Recurrent Neural Networks (RNN), Graph Neural Networks (GNN), autoencoders, Generative Adversarial Networks (GAN), and Restricted Boltzmann Machines (RBM). The paper also discusses evaluation protocols, various publicly available auxiliary information, and data features. Furthermore, the survey concludes with a discussion of the challenges and future research opportunities in enhancing collaborative filtering systems with deep learning.

## 1. INTRODUCTION

Recommender system is a technology that uses user data to provide personalized recommendations, and is widely used in the fields of e-commerce, social media and content recommendation. Its core goal is to recommend content that may be of interest to users by analyzing their historical behaviors and preferences, thereby increasing user satisfaction and platform revenue. The generation of recommendation lists is based on user preferences, item characteristics, historical user-item interactions, and other additional information (e.g., temporal and spatial information) [77]. Current research directions in recommender systems are divided into three main categories: collaborative filtering (CF)-based recommender systems, content filtering (CBF)-based recommender systems, and hybrid recommender systems [78]. CF recommender systems suggest items preferred by users with similar characteristics, while CBF recommender systems try to match items based on the user's previously preferred content. Hybrid recommender systems are a combination of CF and CBF, combining the advantages of both.

Collaborative filtering is one of the most commonly used methods in recommender systems and is based on the principle of learning users' behavioral preferences through their historical interactions with items (clicks, views, ratings, etc.) [84] and predicting their behavioral or latent preferences for items they have not yet interacted with. The learning process of a CF model can be broken down into three main components, including the interaction encoder, the loss function, and the positive (i.e., implicit) feedback that is used when only positive feedback is available negative sampling strategy. Since traditional collaborative filtering techniques learn a user's potential interest in an item by constructing a user-item interaction graph based on a low-dimensional embedding representation, this approach is unable to accurately capture the complexity in user-item interactions, and most of the existing research focuses on designing a more robust interaction encoder to capture the synergistic signals between the user and the item.

In Recent years, deep learning has been gradually applied to recommender systems due to its excellent performance in processing complex data patterns and learning high-dimensional feature representations [85]. Deep neural networks are able to effectively capture nonlinear and complex relationships between users and items, and can encode more complex abstract concepts in higher-level data representations. The popularity of this technique has inspired a series of studies applying various neural network architectures to collaborative filtering (CF), including Multilayer Perceptrons (MLPs), Convolutional Neural Networks (CNNs), Recurrent Neural Networks (RNNs), Graph Neural Networks (GNNs), autoencoders, Generative Adversarial Networks (GANs), and Restricted Boltzmann machines (RBMs).

**Differences between this survey and existing ones.** The main contribution of this paper lies in its comprehensive analysis and categorization of the applications of deep neural networks (DNNs) in collaborative filtering (CF) recommendation systems. Although existing surveys touch upon related topics, they often lack in-depth coverage of the intersection between CF and deep learning. For instance, Zhou et al. [85] introduced recommender systems based on deep learning, but their scope covered content-based recommendation systems rather than focusing on collaborative filtering. In addition, Gao et al. [86] and Wu et al. [87] introduced the applications of graph neural networks in recommendation systems, while Patel et al. [88] introduced CNN-based recommender systems. These surveys only involved the application of a single deep neural network model. While they provide in-depth discussions on individual models, they lack a comprehensive analysis of multiple models and their integration within collaborative filtering. In comparison, this survey offers a structured exploration that bridges the gap about survey between collaborative filtering and deep neural networks. It not only elucidates the foundational concepts underlying collaborative filtering and DNNs but also provides a detailed examination of how DNNs enhance CF systems. The paper systematically categorizes and evaluates various DNN models tailored for CF.

**Approach to papers collection.** In this study, we used Google Scholar as the primary search engine to collect approximately 80 relevant papers published within the last five years (from 2020 to 2024). The collected papers were sourced from reputable publishers such as ACM, Springer, ScienceDirect, and IEEE journals. Figure 1 presents statistical data on the distribution of articles by publisher, while Figure 2 illustrates the percentage distribution by year.

**The structure of this paper is as follows.** Section 2 introduces the background of collaborative filtering systems and deep neural networks. Section 3 first introduces the deep neural network model architecture in collaborative filtering, and then provides a detailed classification framework introducing state-of-the-art techniques. Section 4 discusses the datasets, evaluation metrics, and applications of the selected literature. The fifth part summarizes the entire text and outlines future research directions.

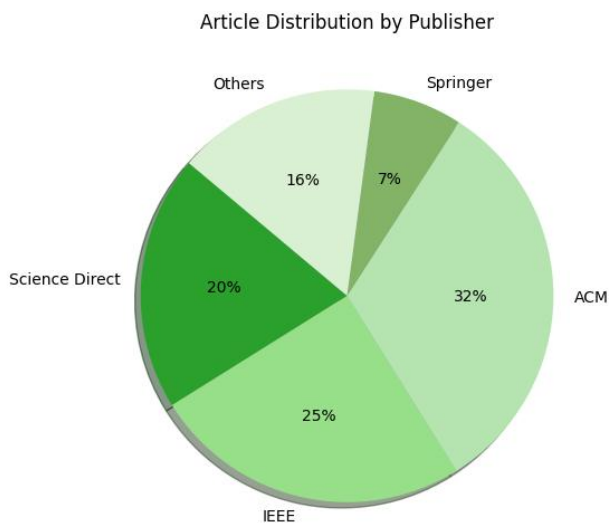

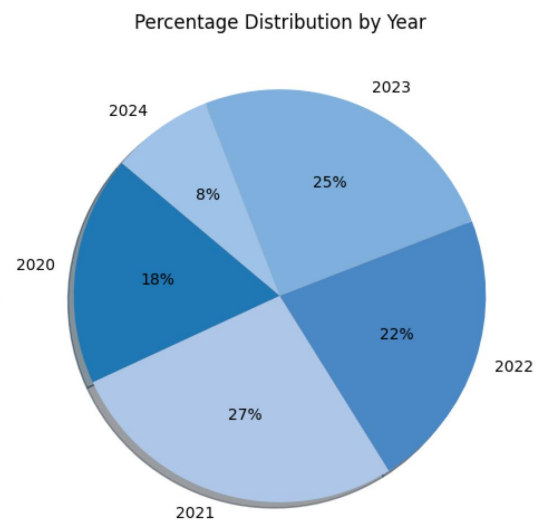

Figure 1: Article distribution by publisher    Figure 2: Percentage distribution by year

## 2. BACKGROUND

In this section, we provide an overview of the two key components that form the foundation of our study: Collaborative Filtering (CF) Recommendation Systems and Deep Neural Networks.

### 2.1 Collaborative Filtering Recommendation Systems

Collaborative Filtering (CF) [78] is a widely used technique in recommendation systems that predicts the interests of a specific user in unknown items by analyzing and comparing user preferences or behavior patterns towards items. This technique assumes that users with similar preferences will have similar interests in items in the future. Collaborative filtering has undergone significant transformation since its inception, marked by advancements from elementary Matrix Factorization (MF) algorithms to sophisticated machine learning techniques.

**Matrix Factorization.** Matrix factorization [89] plays a central role in transforming the sparse user-item interaction matrix into two lower-dimensional latent feature matrices (see Figure 3), where rows represent users and columns represent items. The elements in the matrix represent user ratings or other forms of interaction with items. This technique uses the matrix to identify similarities between users or items and, based on these similarities, predicts items that a user may be interested in. The calculation of similarity typically uses the cosine similarity formula:

$$similarity\ (u,i) = cos\ (\theta) = \frac{r_u \cdot r_i}{\|r_u\| \cdot \|r_i\|},\qquad(1)$$

where and $r_u$ and $r_i$ represent the rating vectors of two users or two items, respectively.

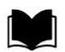

Figure 3: User-item Interaction Matrix

**Challenges with Implicit Feedback.** In dealing with implicit feedback, which is inherently binary (represented by 1s and 0s), traditional matrix factorization reveals a significant drawback. Here, a '1' in the user-item interaction matrix $Y$ indicates that an interaction between a user and an item has been observed, implying that the user has noticed and interacted with the item. Conversely, a '0' does not necessarily signify disinterest but might instead suggest that the user is simply unaware of the item [90]. This dichotomy presents a challenge in learning from implicit data, as it provides only noisy signals about users' preferences. While observed entries at least reflect users' interest in items, the unobserved entries could just be missing data and there is a natural scarcity of negative feedback. The recommendation problem with implicit feedback is formulated as the problem of estimating the scores of unobserved entries in $Y$, which are used for ranking the items. Model-based approaches assume that data can be generated (or described) by an underlying model. Formally, they can be abstracted as learning $y_{ui} = f(u,i\,|\,\Theta)$, where $y_{ui}$ denotes the prediction of user $u$ interacting with item $i$, and $\Theta$ represents the model parameters.

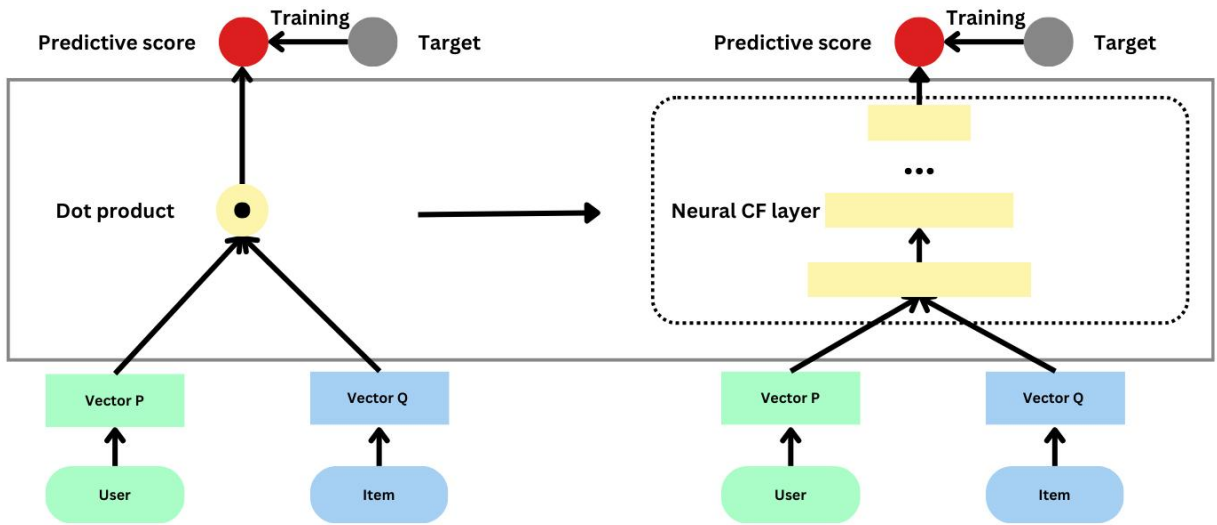

Figure 4: From traditional matrix factorization to neural collaborative filtering

**Neural Collaborative Filtering (NCF)** [90]**.** Figure 4 shows the development of collaborative filtering from dot product methods to neural network models. Traditional dot product methods calculate ratings through the inner product of user and item vectors, while modern neural network models capture the nonlinear relationships between users and items using more complex functions and neural network architectures. The formulation of NCF is given by:

$$r^{\wedge}_{ui} = \sigma(h^T(f(P_u, Q_i))), \tag{2}$$

where $P_u$ and $Q_i$ represent the latent vectors for user u and item i, derived from embedding layers; $f$ denotes the neural network layers processing these vectors; $h$ is the weight vector of the output layer; and σ is an activation function, ensuring outputs are within a probability range. Figure 2 shows the architecture diagram of NCF. By leveraging the depth and flexibility of neural networks, NCF not only effectively addresses the challenges of implicit feedback found in MF but also represents the transition from traditional techniques to more advanced deep learning technologies in collaborative filtering applications [91]. Although NCF addresses the linear limitations of traditional collaborative filtering by incorporating neural network architectures, its performance can still be affected by extremely sparse data. Moreover, while the NCF model may perform well on training data, it lacks the ability to generalize effectively to new users or new items.

## 2.2 Deep Neural Networks

Deep Neural Networks (DNNs) are a subfield of machine learning focused on deep learning, achieving high levels of nonlinear mapping of data through the construction of multilayered network structures, thereby learning complex feature representations and abstract concepts [92]. These networks consist of multiple layers, each containing a large number of neurons, with each neuron functioning as a node performing mathematical operations. As input data is processed through these layers, increasingly abstract features are extracted. A key advantage of deep neural networks is their hierarchical structure [93], which allows them to automatically learn useful features from data without the need for manually designed feature extraction algorithms. In machine learning tasks, deep neural networks are broadly applied to two major types of learning: supervised and unsupervised learning [94]. Supervised learning is one of the most common paradigms in machine learning, where the model is trained using a set of labeled training data with the aim of learning how to map inputs to the correct outputs. In this setup, each training sample is a pair consisting of input features and a corresponding output label. Applications of deep neural networks in this domain include image recognition, speech recognition, and text classification, where the network needs to accurately predict the category labels or continuous output values of the input data. Unsupervised learning, in contrast to supervised learning, does not rely on labeled training samples. This type of learning aims to discover the intrinsic structure and patterns within data, such as through cluster analysis or anomaly detection. Examples of deep neural networks in unsupervised learning include autoencoders and Generative Adversarial Networks (GANs). Autoencoders are used to learn compressed representations of data, while GANs are utilized to generate new data samples that are similar to real data. Through the implementation of these multilayered nonlinear transformations, deep neural networks demonstrate exceptional efficacy in handling complex machine learning tasks. Table 1 illustrates a series of architectural frameworks that are particularly relevant to this survey.

Table 1: Deep neural network model classification

| Deep Neural Network Architecture | Training Type | Description | Publication |
|---|---|---|---|
| Multilayer Perceptron (MLP) | Supervised | Utilizes multiple layers to model complex relationships, enhancing predictive accuracy | [1-7] |
| Convolutional Neural Network (CNN) | Supervised | Uses convolution operations, capturing local features effectively | [8-16], [79-83] |
| Recurrent Neural Network (RNN) | Supervised | Processes sequential data, capturing time-dependent patterns | [17-26] |
| Graph Neural Network (GNN) | Supervised & Unsupervised | Leverages graph structures to model relationships directly | [27-46] |
| Autoencoder | Unsupervised | Uses an encoder-decoder structure to learn dense, compact representations | [47-68], [3] |
| Generative Adversarial Network (GAN) | Unsupervised | Comprises generator and discriminator competing against each other | [69-72] |
| Restricted Boltzmann Machine (RBM) | Unsupervised | A stochastic neural network useful for dimensionality reduction and feature learning | [73-76] |

## 3. NEURAL COLLABORATIVE FILTERING RECOMMENDER SYSTEMS

In this section, we delve into the core and advanced neural network models that have revolutionized collaborative filtering recommender systems. We start with fundamental models, such as Multilayer Perceptrons (MLPs), which form the backbone of neural collaborative filtering. Then, we explore more advanced techniques that represent the state-of-the-art in the field, highlighting their unique approaches and superior performance in capturing user-item interactions.

## 3.1 Fundamental Neural Network Models: Multilayer Perceptrons

Multilayer Perceptrons (MLPs) are a fundamental neural network architecture extensively used in collaborative filtering. Figure 5 shows a sampe of Multilayer Perceptron architecture. MLPs consist of multiple layers of neurons, each layer fully connected to the next, enabling the model to learn complex, non-linear representations of the data. In this context, the interaction between users and items is modeled by concatenating their embedding vectors, $P_u$ and $Q_i$, as $x_{ui} = [P_u, Q_i]$. This concatenated vector is passed through several hidden layers with non-linear activation functions, typically ReLU, as follows: $h_1 = ReLU(W_1 X_{ui} + b_1)$, $h_2 = ReLU(W_2 h1 + b_2)$, ..., $h_L = ReLU(W_L h_{L-1} + b_L)$. Finally, the output layer applies a linear transformation followed by a sigmoid function to predict the rating: $r\hat{}_{ui} = \sigma(W_o h_L + b_o)$. This structure allows MLPs to capture complex patterns and non-linear relationships between users and items, leading to more accurate and personalized recommendations compared to traditional linear methods.

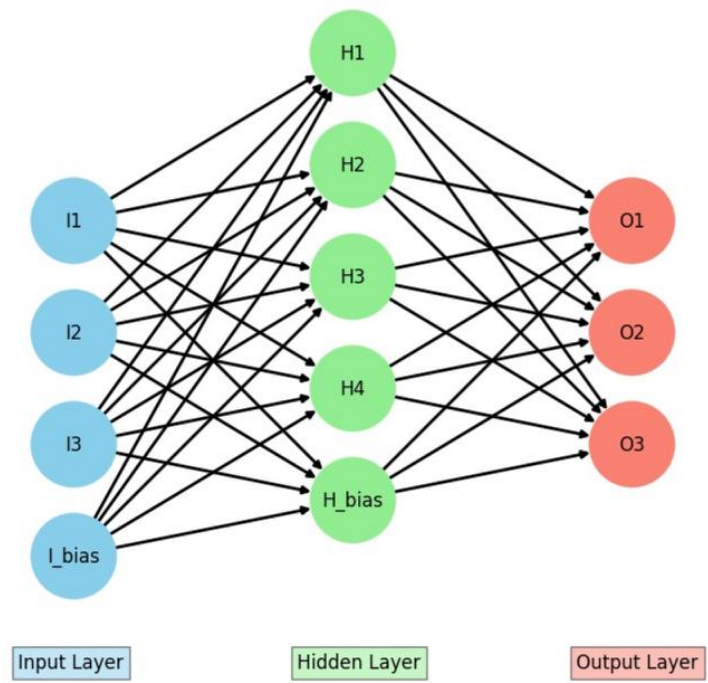

Figure 5: Multilayer Perceptron architecture

In the latest research on collaborative filtering recommendation systems [1-7], several teams have proposed innovative models that utilize neural networks and advanced learning techniques to improve the accuracy and interpretability of recommendations. These studies involve multi-layer perceptrons (MLP), federated learning, and other advanced feature extraction techniques aimed at addressing issues of data sparsity, privacy protection, and system performance.

**Interpretability**. Pugoy and Kao [2] introduced the EscoFilt model, which integrates BERT, K-Means clustering, and MLP for extractive summarization-based collaborative filtering. This model enhances rating prediction accuracy and interpretability by combining advanced feature extraction and clustering techniques with MLP in processing user review data.

**Data sparsity**. Addressing data sparsity issues, several research groups have employed innovative MLP models to optimize data usage. Lin et al. [3] proposed the NCF-MS model, integrating stacked denoising autoencoders (SDAE) and MLP with a cloud-edge collaborative computing architecture to merge multi-source data and address data sparsity. Kim and Lim [4] developed a deep neural collaborative filtering model for e-book service recommendations, utilizing an MLP structure with Bayesian optimization for hyperparameter tuning. Farhan Ullah et al. [7] introduced the Deep Edu model for educational service recommendations, using embedding layers and MLP to effectively learn nonlinear relationships between features.

**Learning explicit ratings and implicit interactions**. Wang et al. [6]'s CEICFNet model integrates MLP to learn explicit ratings and implicit interactions in a cross-domain setting, using domain-shared and domain-specific networks to learn the latent factors of users and items. Yu et al. [5] introduced the CFFNN model, which combines MLP with a cross-feature fusion mechanism to address the relationship between user preferences and item features. This model employs a self-attention mechanism to accurately extract and merge user and item features.

**Protecting user privacy**. Perifanis and Efraimidis [1] proposed the Federated Neural Collaborative Filtering (FedNCF) model, which combines federated learning and neural collaborative filtering to enhance recommendation performance and protect user privacy.

**3.2 Advanced Neural Network Techniques: State-of-the-Art**

In this subsection, recent advancements in neural network techniques have significantly improved collaborative filtering methods, incorporating the below state-of-the-art architectures to address complex challenges.

**3.2.1 Convolutional Neural Networks**

Convolutional Neural Networks (CNNs) are widely used in collaborative filtering due to their powerful feature extraction capabilities. The process begins with an input image representing user-item interactions. The convolutional layer applies convolution operations to extract relevant features from the input data. Following this, pooling layers reduce the dimensionality of the feature maps, retaining the most critical information. The pooled feature maps are then flattened into a vector that serves as input to the fully connected layers. These layers perform the final ranking and output the predicted results. The figure below illustrates the architecture of a CNN in the context of collaborative filtering.

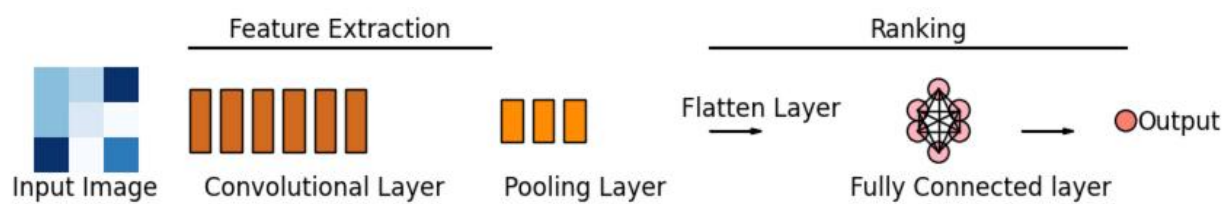

Figure 6: Convolutional Neural Network architecture

**Enhancing Co-occurrence Pattern Recognition.** Recent studies have employed CNNs to enhance collaborative filtering techniques by capturing co-occurrence patterns, leveraging hybrid methods, and improving adversarial robustness and contextual features. Chen et al. [8] introduced the CoCNN model, which leverages co-occurrence patterns within user-item and item-item relationships to enhance recommendation performance. Similarly, Lin et al. [9] proposed the COMET model, which constructs embedding maps from user and item interactions and applies CNNs with varying kernel sizes to extract interaction features. This approach allows COMET to effectively model and leverage complex user-item interactions.

**Addressing the challenge of adversarial attacks.** Gao et al. [10] proposed the Adversarial Neural Collaborative Filtering model with Embedding Dimension Correlations (ANCF). This model employs Adversarial Personalized Ranking (APR) to enhance robustness and uses the outer product to learn pairwise embedding correlations. Bhuvaneshwari et al. [11] introduced a Top-N recommendation system that leverages explicit feedback and outer product-based residual convolutional neural networks (OPBR-CNN). This model utilizes an explicit user-item sparse rating matrix and residual connections to capture complex user-item interaction signals.

**Incorporating user reviews and contextual features.** Dezfouli et al. [12] proposed the MatchPyramid Recommender System (MPRS), which leverages user reviews by framing the recommendation problem as a text matching task. The system constructs a matching matrix from user and item review texts, which is then processed by CNNs to compute matching scores for user-item pairs. Alrashidi et al. [13] developed a social recommender system (SRSCNN) that integrates CNNs with tagging and contextual features. This model captures user and item factors effectively by combining features from item titles and tags, thereby enhancing recommendation accuracy.

**Hybrid methods that integrate various feedback types.** Drammeh and Li [14] enhanced neural collaborative filtering with hybrid feature selection, combining global and local item correlations using pointwise convolution and average pooling. This method captures richer latent signals and prevents overfitting by optimizing network weights with Generalized Matrix Factorization (GMF). Gurav et al. [15] proposed the IA-CNN architecture, which integrates behavioral features into dense vectors using a heatmap matrix. This approach uses convolutional layers to extract new features from user and product word embeddings, capturing complex interactions. Li and Xia [16] introduced the CMAMG_ALSCF model, combining ALS collaborative filtering with deep learning to process web data, reducing noise and normalizing inputs. This system uses CNNs to classify data based on user evaluations and interests, improving movie recommendation accuracy.

### 3.2.2 Recurrent Neural Networks

Figure 7 illustrates the architecture of a Recurrent Neural Network. RNNs are particularly effective for modeling user interactions over time, allowing for more accurate predictions of future user preferences. An RNN processes sequences of data by maintaining a hidden state that captures information from previous time steps. The hidden state at time step $t$ is computed as:

$$ht = \sigma(W_h h_{t-1} + W_x x_t + b_h), \qquad\qquad (3)$$

where $W_h$ and $W_x$ are weight matrices, $b_h$ is a bias vector, $x_t$ is the input at time step $t$, and $\sigma$ is an activation function, typically tanh or ReLU.

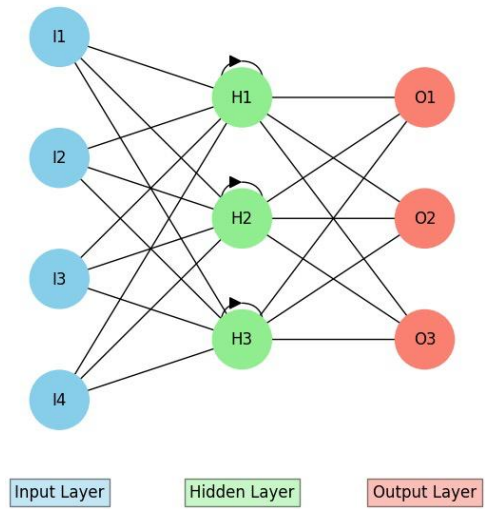

Figure 7: Recurrent Neural Network architecture

**Models Based on LSTM.** Several studies have enhanced recommendation systems using Long Short-Term Memory (LSTM) networks. Manotumruksa et al. [17] introduced the CRCF model, combining LSTM and contextual attention to capture short-term user preferences, considering time and geographical factors. Do and Nguyen [18] developed a semantic-enhanced NCF model using LSTM and knowledge graphs for improved movie recommendations. Fu Chen et al. [25] proposed the SDCF model, leveraging LSTM and self-attention to model learning outcomes in assessments. Karabila et al. [26] combined sentiment analysis and collaborative filtering using Bi-LSTM for personalized e-commerce recommendations. Ebrahimian and Kashef [22] used CNNs and RNNs to detect shilling attacks in recommendation systems. Li et al. [23] integrated LSTM to predict student mastery levels in educational recommendations, optimizing collaborative filtering and cognitive diagnosis.

**Models Based on GRU.** Xia et al. [21] proposed the AGAMF model, which uses attention-based Gated Recurrent Units (GRU) and adversarial learning to improve recommendation performance. GRU extracts contextual relationships from user information, while adversarial learning enhances model robustness. Liang et al. [24] introduced RNCF, a method for Quality of Service (QoS) prediction in the Internet of Vehicles (IoV). This model employs a multi-layer GRU to capture dynamic environmental and network conditions, improving QoS prediction accuracy.

**Models Integrating RNN with Other Techniques.** Chung-Ming Huang and Chen-Yi Wu [19] proposed a POI recommendation method for Mobile Digital Cultural Heritage (M-DCH) using RNNs and user collaborative filtering to analyze user behaviors and preferences. This method achieves high precision, recall, and diversity in recommendations without using a rating mechanism by analyzing historical user behavior and leveraging collaborative filtering of similar users. Ibrahim et al. [20] introduced the HNCF model, integrating hierarchical user and product attention, deep collaborative filtering, and a neural sentiment classifier. This model uses RNNs and MLPs to deeply fuse user and product features, enhancing recommendation performance.

### 3.2.3 Graph Neural Networks

Graph Neural Networks (GNNs) are highly effective for collaborative filtering tasks due to their ability to model relationships in graph-structured data. The following figure illustrates the process of using GNNs in collaborative filtering.

**Graph Construction**. Starting with a user-item interaction matrix, we construct a bipartite graph where nodes represent users and items, and edges represent interactions.

$$\text{Data Matrix} = \begin{bmatrix} 2 & 1 \\ 1 & 0 \end{bmatrix}$$

This matrix is converted into a graph with nodes for users and items, and edges reflecting the interactions.

**Graph Neural Network.** A GNN is then used to learn node embeddings. Each node updates its embedding by aggregating information from its neighbors. The rule is:

$$h_v^{(k)} = \sigma\left( \sum_{u \in \mathcal{N}(v)} W^{(k)} h_u^{(k-1)} \right)$$

(4)

where $h_v{}^{(k)}$ is the embedding of node $vv$ at layer $kk$, $N_{(v)}$ is the set of neighbors of $v$, $W_{(k)}$ is a weight matrix, and σ is an activation function.

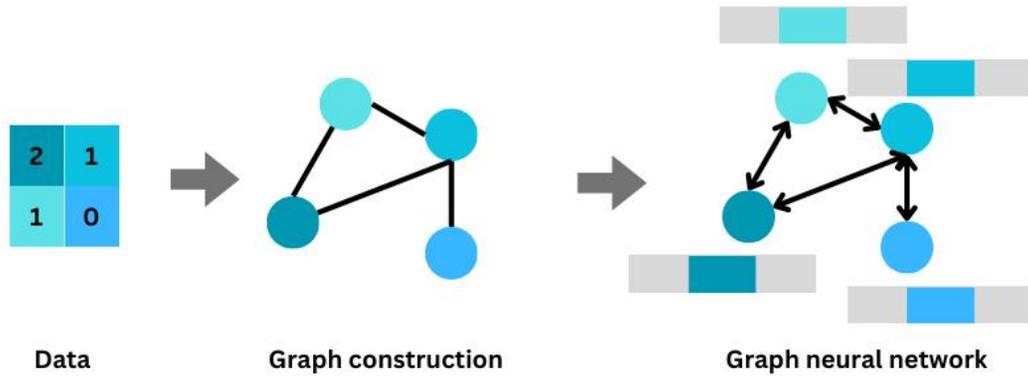

Figure 8: Graph Neural Network architecture

**Contrastive Learning.** Contrastive learning enhances model representation by maximizing similarity from different perspectives. Li et al. [29] proposed the MD-GCCF model, using multi-view deep graph contrastive learning to address over-smoothing in GNNs and enhance feature diversity. Lin et al.'s NCL [27] model improves recommendation quality in sparse data environments by introducing structured and semantic neighborhood concepts. Xia et al.'s HCCF framework [30] and Ren et al.'s DCCF model [40] enhance user and item representations and disentangle intents using hypergraph contrastive learning and self-supervised learning, respectively.

**High-Order Relationship Capturing.** Capturing high-order relationships is crucial in collaborative filtering. Dynamic graph models like DGCF by Xiaohan Li et al. [32] and DMGCF by Tang et al. [28] integrate temporal and collaborative information through dynamic graphs and dynamically generate user and item graphs, achieving more efficient recommendation systems.

**Information Fusion.** Information fusion plays a significant role in enriching the contextual information of recommendation systems. Peng et al.'s KGCFRec model [33] enhances recommendation system performance and robustness by integrating knowledge graph and collaborative filtering information.

**Denoising.** Denoising techniques enhance recommendation performance in noisy data environments. Tian et al.'s RGCF model [42] reduces noisy interaction impact on GNN representation learning with hard and soft denoising strategies. Xia et al.'s SimRec model [45] combines knowledge distillation and contrastive learning to address over-smoothing and noise, maintaining system diversity.

**Interpretability.** Enhancing the interpretability of recommendation systems aids in user trust and system optimization. Wang et al.'s DGCF model [35] improves interpretability by dividing user and item representations into independent components and updating intention-aware graphs iteratively. Wenqi Fan et al. [31] proposed the GTN method, addressing non-adaptive propagation and unreliable interactions in GNN models using trend filtering. Sangeetha et al. [34] introduced NGCF, leveraging high-order connectivity for encoding collaborative signals on a user-item bipartite graph. Chen Li et al. [36] developed SHCF, combining sequential patterns with high-order heterogeneous collaborative signals in a heterogeneous information network. Xia et al. [37] introduced SimRec, transferring knowledge from a teacher GNN to a lightweight student network to preserve global signals and address oversmoothing. Su et al. [38] proposed GMCF, which models intra- and cross-interactions through a graph matching structure. Wang et al. [39] developed ADAPT, an adaptive graph pre-training framework for localized collaborative filtering to tackle data sparsity. Liu et al. [41] created a GNN-based attack method to increase product exposure through gradient optimization. Chen et al. [43] introduced GDSRec, a decentralized collaborative filtering model reweighting social connections based on user preference similarity. Hu et al. [44] proposed MGDCF, applying a distance-based Markov process to incorporate high-order neighbor information in GNN-based collaborative filtering.

### 3.2.4 Autoencoders

Autoencoders are a type of neural network designed to learn efficient codings of input data. They play a crucial role in collaborative filtering by capturing the underlying patterns in user-item interactions. The architecture of an autoencoder consists of two main parts: the encoder and the decoder [51]. The encoder compresses the input data into a latent space representation, often referred to as the "code". This process involves multiple layers that gradually reduce the dimensionality of the data, retaining only the most essential features. The central layer, or the "code" layer, holds the compressed representation of the input data. The decoder then reconstructs the input data from the code. This process involves multiple layers that gradually increase the dimensionality of the data back to its original form. The goal is to achieve an output that closely resembles the original input. In the context of collaborative filtering, autoencoders are used to learn latent factors representing users and items. These latent factors can then be utilized to predict user preferences and generate recommendations.

The Figure 9 illustrates the architecture of an autoencoder. The left side shows the input data being compressed through the encoder layers into the code layer. The right side shows the decoder reconstructing the input data from the code.

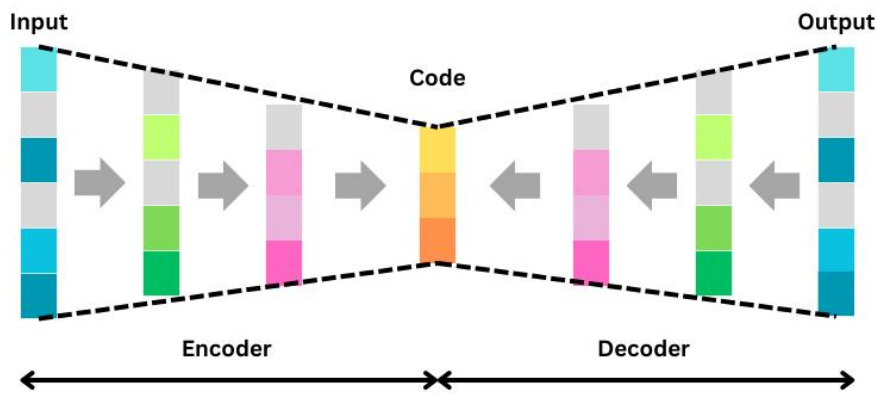

Figure 9: Autoencoder architecture

**Privacy Protection.** Zhang et al. [47] proposed the N3S model, a federated autoencoder framework that combines hybrid negative sampling and secret sharing to protect user data privacy and reduce communication overhead. Polato et al. [66] introduced the Federated Variational Autoencoder (FedVAE) model, which enhances collaborative filtering performance through distributed computation while safeguarding privacy.

**High-Order Relations and Data Augmentation.** Nguyen et al. [48] proposed the HACF framework, using a graph autoencoder to enhance collaborative filtering by integrating high-order connections and data augmentation. Noshad et al. [50] combined mutual information and autoencoders to generate user rating vectors and calculate user similarity. Zheng et al. [64]

introduced the UIAE model, integrating user and item interactions with autoencoders. Xia et al. [53] proposed CRANet, combining reflection reception and information fusion autoencoders to capture implicit user preferences. Bobadilla et al. [65] developed a deep variational model combining VAE and DeepMF for robust latent spaces. Alharbe et al. [49] introduced UI2vec, using word embedding techniques for recommendations. Zeng et al. [61] proposed NCAR, integrating implicit trust relationships and user-item interactions.

**Applications of Variational Autoencoders.** Shenbin et al. [51] introduced RecVAE, a Variational Autoencoder (VAE) for implicit feedback Top-N recommendations, with innovations like a composite prior distribution and user-specific $\beta$ settings, outperforming models like Mult-VAE and RaCT. Zhong and Zhang [62] proposed the Wasserstein Autoencoder (aWAE) for collaborative filtering, using L1 regularization and a new loss function to learn sparse low-rank representations and minimize reconstruction errors. Truong et al. [63] developed the Bilateral Variational Autoencoder (BiVAE) for binary user-item interaction data, combining generative and inference models to capture latent representations and alleviate posterior collapse. Zhong et al. [67] introduced the Gaussian Copula Variational Autoencoder (GCVAE), using the Copula method to better capture dependencies between latent variables and a new reparameterization technique for improved sampling. Carraro et al. [68] proposed a Conditional Variational Autoencoder (C-VAE) to improve interpretability, visualizing latent spaces and clustering users by movie genres for recommendations and profiling, demonstrating effective genre association learning on the MovieLens dataset.

**Efficiency and Scalability.** Chen et al. [52] introduced FastVAE, a fast variational autoencoder using an inverted multi-index for collaborative filtering, achieving efficient sampling in sub-linear or constant time and enhancing sampling quality. Liu et al. [54] developed AutoSeqRec, an efficient sequential recommendation model using multi-information encoders and decoders to capture user preferences. AutoSeqRec updates only the input matrix for incremental recommendations, improving efficiency and accuracy. Vančura et al. [55, 56] proposed ELSA, a scalable linear shallow autoencoder with low-rank plus sparse structures, reducing memory consumption and computation time for large-scale interaction matrices. Spišák et al. [57] introduced sansa, an asymmetric approximate autoencoder for collaborative filtering, which scales to millions of items with low memory requirements and reduced training time. Liu and Wang [58] proposed CFDA, a dual autoencoder system minimizing training data bias to better learn hidden representations of users and items.

**Integration of Social and Contextual Information.** Tahmasebi et al. [59] proposed a social movie recommendation system using deep autoencoder networks and Twitter data. This method combines collaborative and content-based filtering while accounting for users' social influence, measured by their Twitter activity. The system was evaluated with data from MovieTweetings and OMDB. Pan et al. [60] developed a model using deep autoencoders to learn social representations for recommendations. Their Sparse Stacked Denoising Autoencoder (SSDAE) addresses data sparsity and imbalance by extracting features from social information through multi-layer neural networks and matrix factorization techniques.

### 3.2.5 Generative Adversarial Networks

Generative Adversarial Networks (GANs) [71] aim to generate new, synthetic data that resembles a given training dataset. In the context of collaborative filtering, GANs can be used to enhance recommendation systems by generating plausible user-item interaction data. A GAN consists of two main components: the generator network and the discriminator network. The generator network creates fake data by transforming random noise into data points that mimic the real data. The discriminator network, on the other hand, evaluates the authenticity of the data, distinguishing between real data from the training set and fake data produced by the generator.

The Figure 10 illustrates the architecture of a GAN. The left side represents the generator network, which takes random noise as input and generates fake data. The right side depicts the discriminator network, which receives both real and fake data and outputs its evaluation of their authenticity.

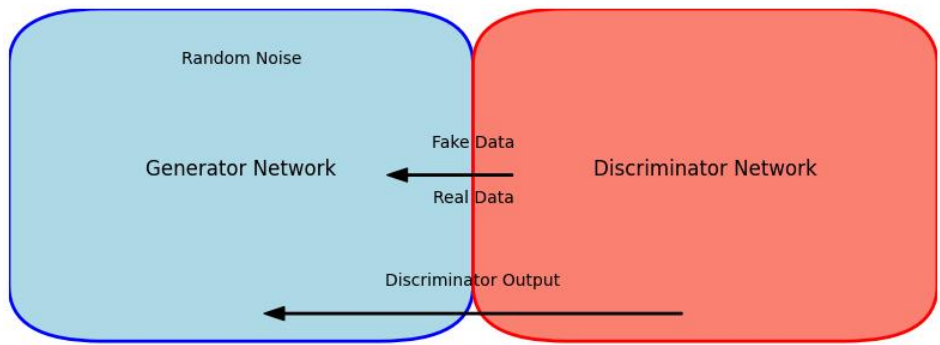

Figure 10: Generative Adversarial Network architecture

Despite the complexity and challenges in GAN training, recent studies have explored their potential in improving collaborative filtering (CF) recommendation systems. Liu et al. [69] proposed CoFiGAN, which enhances recommendation performance through adversarial training by generating positive and negative samples to improve user behavior modeling. Bobadilla et al. [70] introduced a GAN-based method for generating synthetic datasets, using dense embeddings for fast and accurate learning. Dervishaj et al. [71] developed GANMF, applying GANs to matrix factorization and using a feature matching technique for stable training. Ding et al. [72] proposed the BiGAN model, combining ForwardGAN and BackwardGAN to leverage user behavior and friend information for better user and item representations.

### 3.2.6 Restricted Boltzmann Machines

Restricted Boltzmann Machines (RBMs) are a type of stochastic neural network that can learn a probability distribution over its set of inputs. An RBM consists of two layers: a visible layer and a hidden layer. The visible layer represents the input data, while the hidden layer captures the latent features. The connections between these layers are undirected and fully connected, meaning every node in the visible layer is connected to every node in the hidden layer, but there are no connections within a layer. The Figure 11 shows the architecture of an RBM.

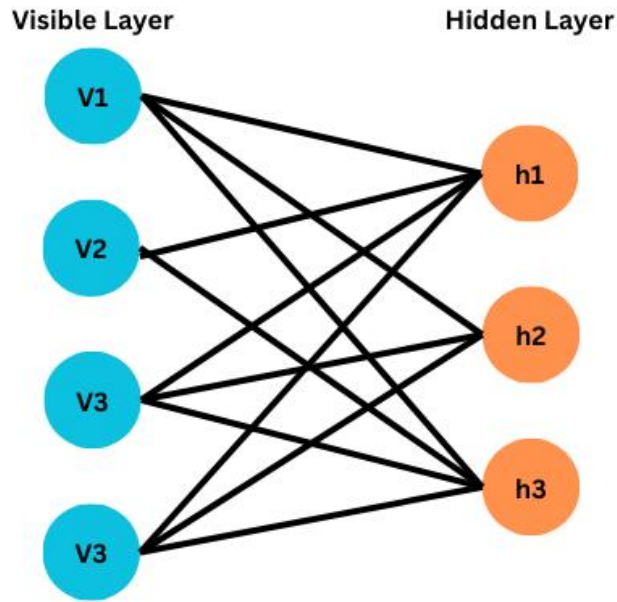

Figure 11: Restricted Boltzmann Machine architecture

Chen et al. [73] proposed the CRBM-IR model, transforming the rating matrix into inspection, positive feedback, and negative feedback matrices to improve recommendation performance using Conditional Restricted Boltzmann Machines. Harshvardhan et al. [74] introduced UBMTR, a time-aware recommendation system that utilizes temporal information and features from user-movie ratings, employing the contrastive divergence algorithm. Kuo et al. [76] presented the DE-CRBM model, combining CRBM with a differential evolution algorithm to optimize parameters and enhance prediction accuracy. Yang et al. [75] proposed a distributed parallel training method using the Horovod framework to accelerate RBM training, significantly reducing training time from 5 hours to under 12 minutes on 64 CPU nodes. This method enhances the feasibility of RBM models in real-world applications requiring frequent retraining.

## 4. DATASETS, EVALUATION METRICS, AND APPLICATIONS

In this section, we provide an overview of the frequently used datasets, evaluation metrics, and the applications and problems addressed by deep neural networks in collaborative filtering recommendation systems.

**4.1 Datasets**

In this section, we outline the most frequently used datasets in collaborative filtering recommendation systems based on an analysis of approximately 80 research papers. The Table 2 lists these datasets along with key statistics such as the number of users, items, ratings, and sparsity, providing valuable resources for training and evaluating recommendation models. Here are brief descriptions of each dataset:

- MovieLens: Collected from the MovieLens website, including MovieLens-1M[1], and MovieLens-10M[2]. These datasets contain user-item rating pairs with timestamps, movie attributes, tags, and user demographic features. Ratings range from 1 to 5, making them widely used benchmarks in collaborative filtering.

- Amazon: Include reviews, product metadata, and links (e.g., also viewed/bought graphs). Sub-datasets like Amazon-Electronics[3] and Amazon-Movies[4] are used to test performance in collaborative filtering and sequential recommendation.

- Pinterest[5]: This dataset includes user interactions with pins on the Pinterest platform, highlighting user engagement and preferences in a social media context. It is often used to test recommendation algorithms in social media settings.

- Yelp[6]: Contains user check-ins and reviews for local businesses. It is frequently updated and used in collaborative filtering and POI (Point of Interest) recommendation tasks, providing insights into local business preferences.

- Gowalla[7]: This check-in dataset includes user locations and social relationships.It is a classical dataset for POI recommendation and is used in both collaborative filtering and sequential recommendation studies.

- LastFM[8]: Features user interactions with music tracks, including user ratings and listening history. This dataset is crucial for developing and testing music recommendation systems, capturing user preferences in the music domain.

Table 2: Statistics of frequently used datasets

| Dataset | #Users | #Items | #Ratings | #Sparsity | Source Link |
|---|---|---|---|---|---|
| MovieLens-1M[1] | 6,040 | 3,706 | 1,000,209 | 95.53 | https://grouplens.org/datasets/movielens/1m/ |
| MovieLens-10M[2] | 69,878 | 10,677 | 10,000,054 | 98.66 | https://grouplens.org/datasets/movielens/10m/ |
| AElectronics[3] | 4,201,696 | 476,002 | 7,824,482 | 99.99 | https://cseweb.ucsd.edu/~jmcauley/datasets.html#amazon_reviews |
| AMovies[4] | 2,121,678 | 207,572 | 7,911,684 | 99.99 | https://cseweb.ucsd.edu/~jmcauley/datasets.html#amazon_reviews |
| Pinterest[5] | 55,187 | 9,916 | 1,500,809 | 99.73 | https://paperswithcode.com/dataset/pinterest |
| Yelp[6] | 1,637,138 | 209,393 | 8,021,122 | 99.97 | https://www.yelp.com/dataset |
| Gowalla[7] | 107,092 | 1,280,969 | 6,442,890 | 99.98 | http://snap.stanford.edu/data/loc-gowalla.html |
| LastFM[8] | 359,347 | 292,385 | 17,559,530 | 99.95 | http://mtg.upf.edu/static/datasets/last.fm/lastfm-dataset-1K.tar.gz |

## 4.2 Evaluation Metrics

In this section, Table 3 shows the evaluation metrics commonly used in the selected papers for assessing the performance of collaborative filtering models. The table below summarizes the metrics and the corresponding papers that reference them. The most frequently used evaluation metrics include Recall@k, NDCG@k, HR@k, RMSE, and MAE. These metrics are essential for evaluating the accuracy and effectiveness of recommendation systems. Recall@k and NDCG@k are often used together to measure the relevance of the top-k recommendations. RMSE and MAE are standard metrics for measuring prediction errors.

Table 3: Statistics of frequently used Metrics

| Metrics | Papers Referenced |
|---|---|
| Recall@k, NDCG@k | [27], [31], [37], [40], [44-46], [51], [52], [55], [57], [62], [63], [67], [82], [83] |
| HR@k, NDCG@k | [1], [3], [5], [6], [8-11], [17], [34], [36], [61], [79] |
| RMSE, MAE | [4], [7], [18], [21], [24], [43], [60] |
| RMSE | [2], [15], [16], [53], [64] |
| Precision@k, Recall@k, F1@k | [19], [49] |
| HR@k | [39], [35] |

## 4.3 Why use deep neural networks in collaborative filtering?

Deep neural networks (DNNs) demonstrate significant advantages in collaborative filtering recommendation systems due to their powerful feature extraction capabilities, high flexibility, ability to handle complex nonlinear relationships, and excellent generalization. DNNs can automatically learn feature representations from vast amounts of data, reducing the reliance on manual feature engineering, and can adapt to various data distributions and diverse user behavior patterns. Through their multi-layer structure, DNNs can capture high-order features and complex interactions within the data, thereby enhancing the accuracy and personalization of recommendations. This section delves into the reasons for incorporating deep neural networks in collaborative filtering and provides a classification of their applications in the field of collaborative filtering recommendation systems, as illustrated in Figure 12.

**Deep neural networks have revolutionized collaborative filtering by addressing several critical system challenges**.

- Privacy Protection: With increasing concerns over data privacy, models like FedNCF [1] employ federated learning to ensure user data remains decentralized and secure, thus protecting user privacy while still enabling effective recommendations.

- Defensive Mechanisms: Deep neural networks enhance the robustness of recommendation systems. Advanced models efficiently detect and mitigate shilling attacks, thereby preserving the integrity and reliability of recommendations.

- Distributed Processing: Leveraging cloud-edge collaboration, models such as NCF-MS [3] efficiently distribute computational tasks, enabling scalable and real-time recommendation processing, which is crucial for handling large-scale data environments.

**Deep neural networks also significantly enhance the performance and optimization of recommendation systems.**

- Cold Start Problem: By incorporating hierarchical and contextual data, models like DeepEdu [7] and HNCF [20] effectively address the cold start problem, facilitating accurate recommendations even for new users and items.

- Data Sparsity: Techniques used in models like FEDNCF [4] and SRSCCNN [13] adeptly manage data sparsity, utilizing sparse data to extract meaningful patterns and improve recommendation accuracy.

- Explicit and Implicit Features: Models such as CEICFNet [6] and MPRS [12] integrate both explicit feedback (like user reviews) and implicit behavior (like browsing history), thereby enriching the recommendation process with comprehensive user insights.

   **Advanced capabilities of deep neural networks further extend their utility in collaborative filtering.**

- Deep Feature Extraction: Models like CoCNN [8]and CFFNN [5] excel in extracting complex co-occurrence relationships and fusing cross-feature information, leading to more accurate and nuanced recommendations.

- Adaptive Propagation: Techniques in models like CARA [17] and Semanticenhanced NCF [18] allow for dynamic adjustment of propagation mechanisms, tailoring the recommendation process to evolving user behaviors and preferences.

- Explainability: Enhancing the transparency and interpretability of recommendations.

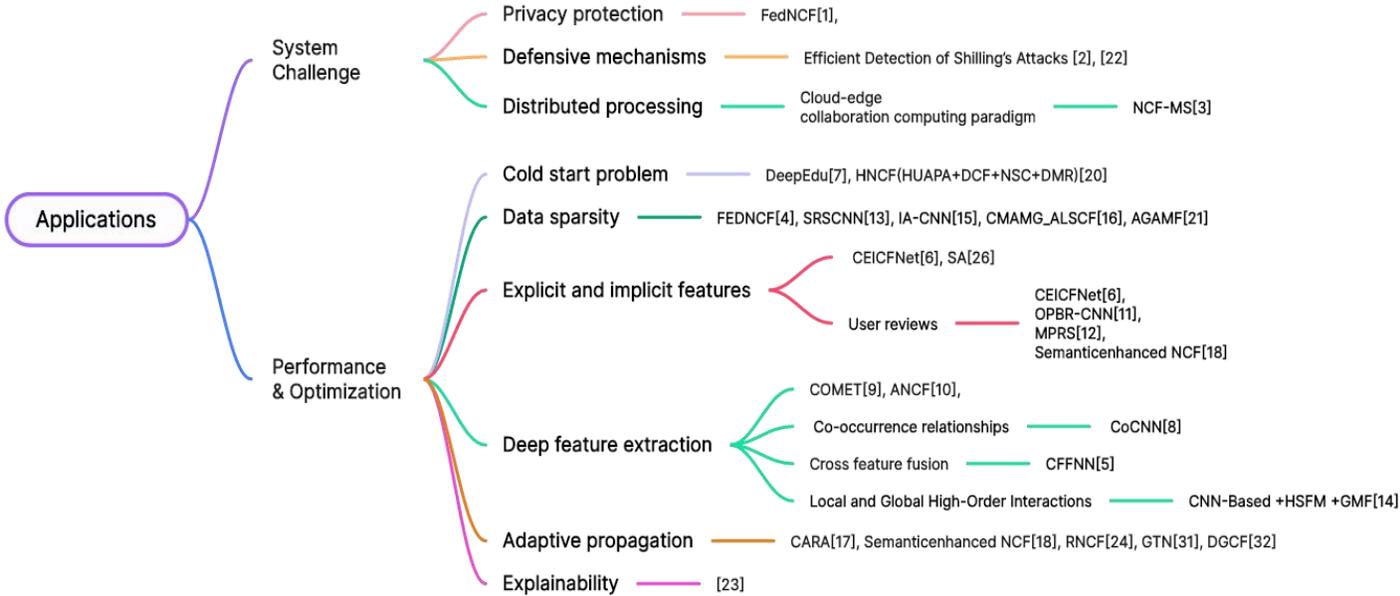

Figure 12: Classifications of Deep Learning Applications in Collaborative Filtering

# 5. CONCLUSION AND FUTURE WORKS

In this survey, we explored the transformative role of deep neural networks (DNNs) in collaborative filtering recommendation systems. DNNs have been proven to be powerful tools for enhancing the accuracy, efficiency, and personalization of recommendations by effectively addressing key challenges such as data sparsity, cold start problems, and the extraction of both explicit and implicit features. Their ability to handle complex nonlinear relationships and capture high-order interactions within the data makes them indispensable in modern recommendation systems. We discussed various DNN architectures and techniques, highlighting their unique contributions and applications. These models have demonstrated significant improvements in capturing user preferences, handling large-scale data, and providing robust and interpretable recommendations. Future research on DNN-based collaborative filtering recommendation systems should focus on enhancing scalability and efficiency to handle larger datasets, improving robustness and security against adversarial attacks, and increasing model interpretability. Additionally, integrating multimodal data, advancing personalization through context-aware models, and optimizing federated learning techniques for privacy preservation are essential areas for further exploration. These advancements will continue to push the boundaries of what recommendation systems can achieve, making them more intelligent, efficient, and user-centric.

## ACKNOWLEDGEMENT

I am deeply grateful to my supervisors, Prof.Shahrul Azman Mohd Noah and Dr. Hafiz Mohd Sarim, for their invaluable guidance, advice, and support.

## REFERENCES

[1]     Perifanis, V., & Efraimidis, P. S. (2022). Federated neural collaborative filtering. *Knowledge-Based Systems, 242*, 108441.

[2]     Pugoy, R. A., & Kao, H. Y. (2021, August). Unsupervised extractive summarization-based representations for accurate and explainable collaborative filtering. *In Proceedings of the 59th Annual*

*Meeting of the Association for Computational Linguistics and the 11th International Joint Conference on Natural Language Processing (Volume 1: Long Papers)* (pp. 2981-2990).

[3]     Lin, W., Zhu, M., Zhou, X., Zhang, R., Zhao, X., Shen, S., & Sun, L. (2023). A deep neural collaborative filtering based service recommendation method with multi-source data for smart cloud-edge collaboration applications. *Tsinghua Science and Technology, 29*(3), 897-910.

[4]     Kim, J. Y., & Lim, C. K. (2023). Feature Extracted Deep Neural Collaborative Filtering for E-Book Service Recommendations. *Applied Sciences*, *13*(11), 6833.

[5]     Yu, R., Ye, D., Wang, Z., Zhang, B., Oguti, A. M., Li, J., ... & Kurdahi, F. (2021). CFFNN: Cross feature fusion neural network for collaborative filtering. *IEEE Transactions on Knowledge and Data Engineering, 34*(10), 4650-4662.

[6]     Wang, C. D., Chen, Y. H., Xi, W. D., Huang, L., & Xie, G. (2021). Cross-domain explicit–implicit-mixed collaborative filtering neural network. *IEEE Transactions on Systems, Man, and Cybernetics: Systems, 52*(11), 6983-6997.

[7]     Ullah, F., Zhang, B., Khan, R. U., Chung, T. S., Attique, M., Khan, K., ... & Jan, S. (2020). Deep edu: a deep neural collaborative filtering for educational services recommendation. *IEEE access*, *8*, 110915-110928.

[8]     Chen, M., Ma, T., & Zhou, X. (2022). CoCNN: Co-occurrence CNN for recommendation. *Expert Systems with Applications*, *195*, 116595.

[9]     Lin, Z., Feng, L., Guo, X., Zhang, Y., Yin, R., Kwoh, C. K., & Xu, C. (2023). Comet: Convolutional dimension interaction for collaborative filtering. *ACM Transactions on Intelligent Systems and Technology, 14*(4), 1-18.

[10]    Gao, Y., Chen, J., Xiao, L., Wang, H., Pan, L., Wen, X., ... & Wu, X. (2023). Adversarial Neural Collaborative Filtering with Embedding Dimension Correlations. *Data Intelligence*, *5*(3), 786-806.

[11]    Bhuvaneshwari, P., Rao, A. N., & Robinson, Y. H. (2023). Top-n recommendation system using explicit feedback and outer product based residual cnn. *Wireless Personal Communications*, *128*(2), 967-983.

[12]    Dezfouli, P. A. B., Momtazi, S., & Dehghan, M. (2021). Deep neural review text interaction for recommendation systems. *Applied Soft Computing*, *100*, 106985.

[13]    Alrashidi, M., Selamat, A., Ibrahim, R., & Fujita, H. (2024). Social Recommender System Based on CNN Incorporating Tagging and Contextual Features. *Journal of Cases on Information Technology (JCIT)*, *26*(1), 1-20.

[14]    Drammeh, B., & Li, H. (2023). Enhancing neural collaborative filtering using hybrid feature selection for recommendation. *PeerJ Computer Science*, *9*, e1456.

[15]    Gurav, Y., Prashanth, S. K., Shaikh, A. A., Ravichand, M., Samarthrao, K. V., & Biradar, V. (2023). Experimental Hybrid Technique for Enhancing the Quality of Personalized Product

Recommendation System using Deep Learning. *International Journal of Intelligent Systems and Applications in Engineering, 11*(4), 376-386.

[16]    Li, N., & Xia, Y. (2024). Movie recommendation based on ALS collaborative filtering recommendation algorithm with deep learning model. *Entertainment Computing*, 100715.

[17]    Manotumruksa, J., Macdonald, C., & Ounis, I. (2020). A contextual recurrent collaborative filtering framework for modelling sequences of venue checkins. *Information Processing & Management, 57*(6), 102092.

[18]    Do, P. M. T., & Nguyen, T. T. S. (2022). Semantic-enhanced neural collaborative filtering models in recommender systems. *Knowledge-Based Systems, 257*, 109934.

[19]    Huang, C. M., & Wu, C. Y. (2021). The point of interest (POI) recommendation for mobile digital culture heritage (M-DCH) based on the behavior analysis using the recurrent neural networks (RNN) and user-collaborative filtering. Journal of Internet Technology, 22(4), 821-833.

[20]    Ibrahim, M., Bajwa, I. S., Sarwar, N., Hajjej, F., & Sakr, H. A. (2023). An intelligent hybrid neural collaborative filtering approach for true recommendations. *IEEE Access*.

[21]    Xia, H., Li, J. J., & Liu, Y. (2020). Collaborative filtering recommendation algorithm based on attention GRU and adversarial learning. *IEEE Access*, 8, 208149-208157.

[22]    Ebrahimian, M., & Kashef, R. (2020, December). Efficient Detection of Shilling's Attacks in Collaborative Filtering Recommendation Systems Using Deep Learning Models. In *2020 IEEE International Conference on Industrial Engineering and Engineering Management (IEEM)* (pp. 460-464). IEEE.

[23]    Li, Z., Hu, H., Xia, Z., Zhang, J., Li, X., Shi, J., ... & Li, X. (2021, July). Exercise recommendation algorithm based on improved collaborative filtering. *In 2021 International Conference on Advanced Learning Technologies* (ICALT) (pp. 47-49). IEEE.

[24]    Liang, T., Chen, M., Yin, Y., Zhou, L., & Ying, H. (2021). Recurrent neural network based collaborative filtering for QoS prediction in IoV. *IEEE Transactions on Intelligent Transportation Systems, 23*(3), 2400-2410.

[25]    Chen, F., Lu, C., Cui, Y., & Gao, Y. (2022). Learning outcome modeling in computer-based assessments for learning: A sequential deep collaborative filtering approach. I*EEE Transactions on Learning Technologies, 16*(2), 243-255.

[26]    Karabila, I., Darraz, N., El-Ansari, A., Alami, N., & El Mallahi, M. (2023). Enhancing collaborative filtering-based recommender system using sentiment analysis. *Future Internet, 15*(7), 235.

[27]    Lin, Z., Tian, C., Hou, Y., & Zhao, W. X. (2022, April). Improving graph collaborative filtering with neighborhood-enriched contrastive learning. *In Proceedings of the ACM web conference 2022* (pp. 2320-2329).

[28]    Tang, H., Zhao, G., Bu, X., & Qian, X. (2021). Dynamic evolution of multi-graph based collaborative filtering for recommendation systems. *Knowledge-Based Systems, 228*, 107251.

[29]    Li, X., Tian, Y., Dong, B., & Ji, S. (2024). MD-GCCF: Multi-view deep graph contrastive learning for collaborative filtering. *Neurocomputing, 590*, 127756.

[30]    Xia, L., Huang, C., Xu, Y., Zhao, J., Yin, D., & Huang, J. (2022, July). Hypergraph contrastive collaborative filtering. *In Proceedings of the 45th International ACM SIGIR conference on research and development in information retrieval* (pp. 70-79).

[31]    Fan, W., Liu, X., Jin, W., Zhao, X., Tang, J., & Li, Q. (2022, July). Graph trend filtering networks for recommendation. *In Proceedings of the 45th international ACM SIGIR conference on research and development in information retrieval* (pp. 112-121).

[32]    Li, X., Zhang, M., Wu, S., Liu, Z., Wang, L., & Philip, S. Y. (2020, November). Dynamic graph collaborative filtering. *In 2020 IEEE international conference on data mining* (ICDM)(pp. 322-331). IEEE.

[33]    Peng, J., Gong, J., Zhou, C., Zang, Q., Fang, X., Yang, K., & Yu, J. (2024). KGCFRec: Improving Collaborative Filtering Recommendation with Knowledge Graph. *Electronics, 13*(10), 1927.

[34]    Sangeetha, M., Manjuladevi, R., Sagana, C., Suruthi, B. S., Thejesvika, S. S., & Vaibhav, R. (2022, March). Predicting personalized recommendations using GNN. *In 2022 6th International Conference on Computing Methodologies and Communication* (ICCMC) (pp. 228-234). IEEE.

[35]    Wang, Y., Li, C., Li, M., Jin, W., Liu, Y., Sun, H., ... & Tang, J. (2022). Localized graph collaborative filtering. I*n Proceedings of the 2022 SIAM International Conference on Data Mining* (SDM) (pp. 540-548). Society for Industrial and Applied Mathematics.

[36]    Li, C., Hu, L., Shi, C., Song, G., & Lu, Y. (2021). Sequence-aware heterogeneous graph neural collaborative filtering. *In Proceedings of the 2021 SIAM International Conference on Data Mining* (SDM) (pp. 64-72). Society for Industrial and Applied Mathematics.

[37]    Xia, L., Huang, C., Shi, J., & Xu, Y. (2023, April). Graph-less collaborative filtering. *In Proceedings of the ACM Web Conference 2023* (pp. 17-27).

[38]    Su, Y., Zhang, R., M. Erfani, S., & Gan, J. (2021, July). Neural graph matching based collaborative filtering. *In Proceedings of the 44th international ACM SIGIR conference on research and development in information retrieval* (pp. 849-858).

[39]    Wang, Y., Li, C., Liu, Z., Li, M., Tang, J., Xie, X., ... & Yu, P. S. (2022). An adaptive graph pre-training framework for localized collaborative filtering. *ACM Transactions on Information Systems, 41*(2), 1-27.

[40]    Ren, X., Xia, L., Zhao, J., Yin, D., & Huang, C. (2023, July). Disentangled contrastive collaborative filtering. In Proceedings of the 46th *International ACM SIGIR Conference on Research and Development in Information Retrieval* (pp. 1137-1146).

[41]    Wang, Y., Liu, Y., & Shen, Z. (2023, June). Revisiting item promotion in GNN-based collaborative filtering: a masked targeted topological attack perspective. *In Proceedings of the AAAI Conference on Artificial Intelligence* (Vol. 37, No. 12, pp. 15206-15214).

[42]    Tian, C., Xie, Y., Li, Y., Yang, N., & Zhao, W. X. (2022, July). Learning to denoise unreliable interactions for graph collaborative filtering. *In Proceedings of the 45th international ACM SIGIR conference on research and development in information retrieval* (pp. 122-132).

[43]    Chen, J., Xin, X., Liang, X., He, X., & Liu, J. (2022). GDSRec: Graph-based decentralized collaborative filtering for social recommendation. *IEEE Transactions on Knowledge and Data Engineering, 35*(5), 4813-4824.

[44]    Hu, J., Hooi, B., Qian, S., Fang, Q., & Xu, C. (2024). Mgdcf: Distance learning via markov graph diffusion for neural collaborative filtering. *IEEE Transactions on Knowledge and Data Engineering*.

[45]    Xia, L., Huang, C., Shi, J., & Xu, Y. (2023, April). Graph-less collaborative filtering. *In Proceedings of the ACM Web Conference 2023* (pp. 17-27).

[46]    Wang, X., Jin, H., Zhang, A., He, X., Xu, T., & Chua, T. S. (2020, July). Disentangled graph collaborative filtering. *In Proceedings of the 43rd international ACM SIGIR conference on research and development in information retrieval* (pp. 1001-1010).

[47]    Zhang, L., Li, G., Yuan, L., Ding, X., & Rong, Q. (2024). HN3S: A Federated AutoEncoder framework for Collaborative Filtering via Hybrid Negative Sampling and Secret Sharing. *Information Processing & Management*, *61*(2), 103580.

[48]    Nguyen, M., Yu, J., Nguyen, T., & Yongchareon, S. (2022). High-order autoencoder with data augmentation for collaborative filtering. *Knowledge-Based Systems*, *240*, 107773.

[49]    Alharbe, N., Rakrouki, M. A., & Aljohani, A. (2023). A collaborative filtering recommendation algorithm based on embedding representation. *Expert Systems with Applications*, *215*, 119380.

[50]    Noshad, Z., Bouyer, A., & Noshad, M. (2021). Mutual information-based recommender system using autoencoder. *Applied Soft Computing*, *109*, 107547.

[51]    Shenbin, I., Alekseev, A., Tutubalina, E., Malykh, V., & Nikolenko, S. I. (2020, January). Recvae: A new variational autoencoder for top-n recommendations with implicit feedback. In *Proceedings of the 13th international conference on web search and data mining* (pp. 528-536).

[52]    Chen, J., Lian, D., Jin, B., Huang, X., Zheng, K., & Chen, E. (2022, April). Fast variational autoencoder with inverted multi-index for collaborative filtering. In *Proceedings of the ACM Web Conference 2022* (pp. 1944-1954).

[53]    Xia, L., Huang, C., Xu, Y., Xu, H., Li, X., & Zhang, W. (2021). Collaborative reflection-augmented autoencoder network for recommender systems. *ACM Transactions on Information Systems (TOIS)*, *40*(1), 1-22.

[54]    Liu, S., Liu, J., Gu, H., Li, D., Lu, T., Zhang, P., & Gu, N. (2023, October). Autoseqrec: Autoencoder for efficient sequential recommendation. In *Proceedings of the 32nd ACM International Conference on Information and Knowledge Management* (pp. 1493-1502).

[55]    Vančura, V., Alves, R., Kasalický, P., & Kordík, P. (2022, September). Scalable linear shallow autoencoder for collaborative filtering. In *Proceedings of the 16th ACM Conference on Recommender Systems* (pp. 604-609).

[56]    Vančura, V. (2023, June). Scalable and Explainable Linear Shallow Autoencoders for Collaborative Filtering from Industrial Perspective. In *Proceedings of the 31st ACM Conference on User Modeling, Adaptation and Personalization* (pp. 290-295).

[57]    Spišák, M., Bartyzal, R., Hoskovec, A., Peska, L., & Tůma, M. (2023, September). Scalable approximate nonsymmetric autoencoder for collaborative filtering. In *Proceedings of the 17th ACM Conference on Recommender Systems* (pp. 763-770).

[58]     Liu, X., & Wang, Z. (2022, July). CFDA: collaborative filtering with dual autoencoder for recommender system. In *2022 International Joint Conference on Neural Networks (IJCNN)* (pp. 1-7). IEEE.

[59]     Tahmasebi, H., Ravanmehr, R., & Mohamadrezaei, R. (2021). Social movie recommender system based on deep autoencoder network using Twitter data. *Neural Computing and Applications*, *33*(5), 1607-1623.

[60]     Pan, Y., He, F., & Yu, H. (2020). Learning social representations with deep autoencoder for recommender system. *World Wide Web*, *23*(4), 2259-2279.

[61]     Zeng, W., Qin, J., & Wei, C. (2021). Neural collaborative autoencoder for recommendation with co-occurrence embedding. *IEEE Access*, *9*, 163316-163324.

[62]     Zhang, X., Zhong, J., & Liu, K. (2021). Wasserstein autoencoders for collaborative filtering. *Neural Computing and Applications*, *33*(7), 2793-2802.

[63]     Truong, Q. T., Salah, A., & Lauw, H. W. (2021, March). Bilateral variational autoencoder for collaborative filtering. In *Proceedings of the 14th ACM International Conference on Web Search and Data Mining* (pp. 292-300).

[64]     Zheng, H., Xing, X., Han, Q., Xin, M., & Niu, Y. (2021, July). UIAE: Collaborative Filtering for User and Item based on Auto-Encoder. In *2021 7th Annual International Conference on Network and Information Systems for Computers (ICNISC)* (pp. 1-6). IEEE.

[65]     Bobadilla, J., Ortega, F., Gutiérrez, A., & González-Prieto, Á. (2023). Deep variational models for collaborative filtering-based recommender systems. *Neural Computing and Applications*, *35*(10), 7817-7831.

[66]     Polato, M. (2021, July). Federated variational autoencoder for collaborative filtering. In *2021 International Joint Conference on Neural Networks (IJCNN)* (pp. 1-8). IEEE.

[67]     Zhong, T., Wang, G., Walker, J., Zhang, K., & Zhou, F. (2021). Variational Autoencoder with Copula for Collaborative Filtering. In *Workshop on Deep Learning Practice for High-Dimensional Sparse Data with KDD.*

[68]     Carraro, T., Polato, M., & Aiolli, F. (2020, July). A look inside the black-box: Towards the interpretability of conditioned variational autoencoder for collaborative filtering. In *Adjunct publication of the 28th ACM conference on user modeling, adaptation and personalization* (pp. 233-236).

[69]     Liu, J., Pan, W., & Ming, Z. (2020). CoFiGAN: Collaborative filtering by generative and discriminative training for one-class recommendation. *Knowledge-Based Systems, 191*, 105255.

[70]     Bobadilla, J., Gutiérrez, A., Yera, R., & Martínez, L. (2023). Creating synthetic datasets for collaborative filtering recommender systems using generative adversarial networks. *Knowledge-Based Systems, 280*, 111016.

[71]     Dervishaj, E., & Cremonesi, P. (2022, April). GAN-based matrix factorization for recommender systems. *In Proceedings of the 37th ACM/SIGAPP Symposium on Applied Computing* (pp. 1373-1381).

[72]    Ding, R., Guo, G., Yan, X., Chen, B., Liu, Z., & He, X. (2020). BiGAN: collaborative filtering with bidirectional generative adversarial networks. *In Proceedings of the 2020 SIAM international conference on data mining* (pp. 82-90). Society for Industrial and Applied Mathematics.

[73]    Chen, Z., Ma, W., Dai, W., Pan, W., & Ming, Z. (2020). Conditional restricted Boltzmann machine for item recommendation. *Neurocomputing, 385*, 269-277.

[74]    Harshvardhan, G. M., Gourisaria, M. K., Rautaray, S. S., & Pandey, M. (2022). UBMTR: Unsupervised Boltzmann machine-based time-aware recommendation system. *Journal of King Saud University-Computer and Information Sciences, 34*(8), 6400-6413.

[75]    Yang, P., Varadharajan, S., Wilson, L. A., Smith, D. D., Lockman III, J. A., Gundecha, V., & Ta, Q. (2020). Parallelized training of restricted boltzmann machines using markov-chain monte carlo methods. *SN Computer Science, 1*(3), 165.

[76]    Kuo, R. J., & Chen, J. T. (2020). An application of differential evolution algorithm-based restricted Boltzmann machine to recommendation systems. Journal of Internet Technology, 21(3), 701-712.

[77]    Li, Z., Zhao, H., Liu, Q., Huang, Z., Mei, T., & Chen, E. (2018, July). Learning from history and present: Next-item recommendation via discriminatively exploiting user behaviors. *In Proceedings of the 24th ACM SIGKDD international conference on knowledge discovery & data mining* (pp. 1734-1743).

[78]    Thorat, P. B., Goudar, R. M., & Barve, S. (2015). Survey on collaborative filtering, content-based filtering and hybrid recommendation system. I*nternational Journal of Computer Applications, 110*(4), 31-36.

[79]    Chen, L., Wu, L., Hong, R., Zhang, K., & Wang, M. (2020, April). Revisiting graph based collaborative filtering: A linear residual graph convolutional network approach. *In Proceedings of the AAAI conference on artificial intelligence*(Vol. 34, No. 01, pp. 27-34).

[80]    Xia, J., Li, D., Gu, H., Lu, T., Zhang, P., & Gu, N. (2021, October). Incremental graph convolutional network for collaborative filtering. *In Proceedings of the 30th ACM International Conference on Information & Knowledge Management* (pp. 2170-2179).

[81]    Mei, D., Huang, N., & Li, X. (2021). Light graph convolutional collaborative filtering with multi-aspect information. *IEEE Access*, 9, 34433-34441.

[82]    He, L., Wang, X., Wang, D., Zou, H., Yin, H., & Xu, G. (2023, February). Simplifying graph-based collaborative filtering for recommendation. *In Proceedings of the Sixteenth ACM International Conference on Web Search and Data Mining* (pp. 60-68).

[83]    Sun, J., Cheng, Z., Zuberi, S., Pérez, F., & Volkovs, M. (2021, April). Hgcf: Hyperbolic graph convolution networks for collaborative filtering. *In Proceedings of the Web Conference 2021* (pp. 593-601).

[84]    Lei, W., He, X., Miao, Y., Wu, Q., Hong, R., Kan, M. Y., & Chua, T. S. (2020, January). Estimation-action-reflection: Towards deep interaction between conversational and recommender systems. *In Proceedings of the 13th International Conference on Web Search and Data Mining* (pp. 304-312).

[85]    Zhou, H., Xiong, F., & Chen, H. (2023). A comprehensive survey of recommender systems based on deep learning. *Applied Sciences, 13*(20), 11378.

[86]    Gao, C., Zheng, Y., Li, N., Li, Y., Qin, Y., Piao, J., ... & Li, Y. (2023). A survey of graph neural networks for recommender systems: Challenges, methods, and directions. *ACM Transactions on Recommender Systems, 1*(1), 1-51.

[87]    Wu, S., Sun, F., Zhang, W., Xie, X., & Cui, B. (2022). Graph neural networks in recommender systems: a survey. *ACM Computing Surveys, 55*(5), 1-37.

[88]    Patel, R., Thakkar, P., & Ukani, V. (2024). CNNRec: Convolutional Neural Network based recommender systems-A survey. *Engineering Applications of Artificial Intelligence, 133*, 108062.

[89]    Rendle, S., Krichene, W., Zhang, L., & Anderson, J. (2020, September). Neural collaborative filtering vs. matrix factorization revisited. *In Proceedings of the 14th ACM Conference on Recommender Systems* (pp. 240-248).

[90]    He, X., Liao, L., Zhang, H., Nie, L., Hu, X., & Chua, T. S. (2017, April). Neural collaborative filtering. *In Proceedings of the 26th international conference on world wide web* (pp. 173-182).

[91]    Wen, H., Liu, X., Yan, C., Jiang, L., Sun, Y., Zhang, J., & Yin, H. (2019, October). Leveraging multiple implicit feedback for personalized recommendation with neural network. *In Proceedings of the 2019 International Conference on Artificial Intelligence and Advanced Manufacturing* (pp. 1-6).

[92]    Hatcher, W. G., & Yu, W. (2018). A survey of deep learning: Platforms, applications and emerging research trends. *IEEE access, 6*, 24411-24432.

*[93]*    Chen, M., Bai, Y., Lee, J. D., Zhao, T., Wang, H., Xiong, C., & Socher, R. (2020). Towards understanding hierarchical learning: Benefits of neural representations. *Advances in Neural Information Processing Systems, 33*, 22134-22145.

[94]    Möller, D. P. (2023). Machine Learning and Deep Learning. I*n Guide to Cybersecurity in Digital Transformation: Trends, Methods, Technologies, Applications and Best Practices* (pp. 347-384). Cham: Springer Nature Switzerland.